\documentclass[10pt,a4paper]{article}
\usepackage{graphicx} % Required for inserting images
\usepackage{amsmath,amssymb,amsthm}
\usepackage{bm}
\usepackage{mathtools}
\usepackage{amsfonts}
\usepackage{booktabs}   % \toprule, \midrule, \bottomrule, \cmidrule
\usepackage{makecell}   % \makecell for two-line headers
\usepackage{graphicx}   % \resizebox
\usepackage{multirow} 
\usepackage{tabularx}
\usepackage{booktabs}
\usepackage{array}
\usepackage{float}
\usepackage{hyperref}
\usepackage{wrapfig}
\usepackage{url}
\usepackage{caption}

\newtheorem{theorem}{Theorem}

\title{Estimation-Aware Reachability}
\author{Himanshu Paudel and Javier Borquez }
\date{}
\usepackage{titling}
\begin{document}

\maketitle
\vspace{-1cm}
\begin{abstract}

Hamilton–Jacobi Reachability computes the states from which a system can be kept safe under worst-case disturbances, but it assumes the state is known exactly, whereas real robots act on noisy estimates whose uncertainty evolves with their motion. We extend Hamilton–Jacobi Reachability to the robot’s belief, estimate and uncertainty, treating unexpected measurements as a bounded adversary and enlarging the unsafe set by the current uncertainty, so that certified states are safe with a prescribed probability. The formulation is compatible with existing solvers. In simulation, the resulting safety filters nearly eliminate the collisions of certainty-equivalent HJ and are less conservative than equally safe uncertainty-padding baselines. The implementation code is available at :\url{https://github.com/HimanshuPaudel/Estimation-Aware-Reachability}.

\end{abstract}
\section{Introduction}
Ensuring safety for autonomous and robotic systems operating under uncertainty has motivated a broad family of set-based safety filter approaches. Some of the popular approaches include Control Barrier Functions (CBFs), \cite{cbf, ames2016control} Zonotope-based methods\cite{zonotope, nonlin_zono_2015,zono_poly_2010}, and Hamilton-Jacobi (HJ)  reachability analysis \cite{hjr,mitchell2004toolbox}. HJ reachability analysis is of particular interest as it provides a general pipeline to compute the maximal safe set for the system, that is, the largest set of states from which the system can be kept safe for a future time horizon while providing the corresponding optimal control policy to avoid that keeps the system in this safe set, without requiring a hand designed safety certificate and naturally handles disturbances. 

%The classical HJR analysis suffers from a fundamental limitation due to the curse of dimensionality, rendering the problem intractable for higher dimensional systems. (true but we are not addressing this directly so it shouldnt have the spotlight)

Classical HJ reachability formulation assumes perfect state observability, an assumption that rarely holds in real hardware setups of autonomous systems, where state estimation is generally used to infer the true state from the noisy measurements. The mismatch between the assumption of full information setting and practical state estimation setting motivates the need for reachability methods that accounts for estimation uncertainty. Doing so, increases the complexity of the underlying problem, further straining classical grid-based HJ reachability solvers that scale poorly with problem size due to the curse of dimensionality. Several lines of work address this scalability challenge such as decomposition methods\cite{decomp} and learning based approaches. Learning based approaches approximate the reachable sets\cite{deepreach, jaime_rlhjr} without explicitly solving the Hamilton-Jacobi equations over the grid, and there has been some recent developments to provide formal assurances to such learning based approaches \cite{fassurance, vnrt}.

We propose an estimation aware reachability framework that augments the system state with the estimator's mean and covariance. This is generally avoided in the literature due to the curse of dimensionality but recent approaches such as DeepReach~\cite{deepreach} make high-dimensional problems tractable. Specifically, we bound the measurement innovation by the instantaneous covariance, and let this bound enter the reachability computation as a worst case disturbance the value function must remain robust to. This yields a value function and safety filter that is aware of and adapts to the system's real-time estimation confidence, rather than treating estimation error as a static, decoupled quantity layered on top of a reachable set computed with the assumption of perfect information of states. %Since the standard target function, commonly defined as a signed distance function, depends only on the physical state and is therefore blind to estimation confidence, we further propose modifications to it that account for the covariance.

We validate this formulation on three case studies: the Air3D pursuit-evasion problem with a continuous-time EKF (CT-EKF), a planar quadrotor with a continuous-time extended $H_\infty$ filter, and a Dubins car whose full state is estimated by a CT-EKF from position measurements of a beacon whose noise grows with distance. In each, the estimate is augmented with the filter's Riccati state: the covariance $\Sigma$ for the CT-EKF, and the $H_\infty$ bound $P$ on the error covariance for the $H_\infty$ filter. We analyze how the resulting Backward Reachable Tube (BRT) changes with the estimator's uncertainty, which provides a concrete testbed for the estimation-aware formulation and a way to visualize the coupling between state uncertainty and the guaranteed safe set.

% This gives a six-dimensional augmented state for Air3D and nine-dimensional ones for the quadrotor and the Dubins car. The three systems cover progressively richer covariance dynamics: in the quadrotor the covariance evolves independently of the state and control, in Air3D it depends on the control, and in the Dubins car it depends on both the position, through the measurement noise, and the heading, through the linearized dynamics, so that heading uncertainty feeds into position uncertainty and the vehicle's path determines how much information it gains.

Our contributions are:
\begin{itemize}
    \item An estimation-aware HJ reachability formulation over a belief-augmented state in which the measurement innovation acts as an adversarial input.
    \item An uncertainty-inflated target set with point-wise probabilistic safety guarantees.
    \item A formulation directly compatible with existing state of the art pipelines for reachable set calculations.
    % \item Three case studies with two estimators: a 6D Air3D pursuit-evasion problem with a CT-EKF, validated against a grid solution; a 9D planar quadrotor with an extended $H_\infty$ filter; and a 9D Dubins car with a CT-EKF over its full state and position-dependent measurement noise. The learned value functions are used as least-restrictive safety filters \cite{hjrsafety} and compared with     filters built from the nominal BRT.
\end{itemize}

The remainder of this paper is organized as follows. Section~\ref{sec:related} reviews related work on reachability under state uncertainty. Section~\ref{sec:background} reviews HJ reachability DeepReach, and the CT-EKF and $H_\infty$ filters. Section~\ref{sec:psetup} formulates the problem and Section~\ref{sec:method} presents the estimation-aware reachability formulation, including the augmented state dynamics, the bound on the measurement innovation via the instantaneous covariance, and the probabilistic guarantee. Section~\ref{sec:case} applies the formulation to the three systems and describes the DeepReach-based training procedure and the safety filter implementation. Finally, Section~\ref{sec:conclusion} concludes and outlines directions for future work.

\section{Related Work}
\label{sec:related}

Several works have extended HJ reachability to relax the assumption of perfect state information, typically by augmenting the reachability state or the closed-loop system with some form of belief. Closest to our setting in HJ reachability literature is  RoVer-CoRe\cite{rovercore} concatenating a controller, observation function, and state estimator into a single closed loop system verifiable with standard reachability tools. Related lines of work use parameter-conditioned reachability to adapt safety assurances online without re-solving the reachability problem \cite{parameter}, formulate HJ reachability over a joint human state and model-parameter belief for robust human motion prediction \cite{human}, solve a belief-space reach-avoid game over an opponent’s type\cite{deception}, or develop planner-agnostic, real-time HJ safety analysis for vehicles in partially observed environments \cite{bajcy}. Others adapt offline value functions online, by local patching \cite{patching} or by a time-varying formulation for spatially varying disturbances \cite{space2time}, but none of these explicitly accounts for state estimation uncertainty. 

% Stochastic reachability methods\cite{sreach} have also been developed to construct probabilistic reachable sets, although they do not generally follow the HJ reachability framework.

In CBF literature some methods handle estimation uncertainty by evaluating the barrier at the state estimate and adding a margin or risk constraint. Belief CBFs \cite{bcbf} are closest to our setting: defining barriers over EKF beliefs and bounding safe-set violation probabilities under measurement updates. Other methods tighten the barrier condition using either a deterministic bound on the estimation error \cite{mr-cbf, rcbf, ofbcbf} or concentration inequalities on the Gaussian estimate \cite{pcbf}. In all of these, the guarantee is certified for a user-specified barrier or backup set and enforced by a reactive safety filter. In contrast, we compute the avoid set itself with HJ reachability over the augmented (state, covariance) space, which yields the safety value function and an optimal safety controller, including against an adversarial disturbance.
\section{Background}
\label{sec:background}
\subsection{Hamilton-Jacobi Reachability}
Consider a system with state $x \in \mathbb{R}^n$, control input $u \in \mathcal{U}$,
and bounded disturbance $d \in \mathcal{D}$, evolving under dynamics
\begin{equation}
    \dot{x} = f(x, u, d).
    \label{eq:dynamics}
\end{equation}

HJ reachability analysis characterizes the set of states from which a system can
be driven into (or kept out of) a target set $\mathcal{L} \subseteq \mathbb{R}^n$,
typically defined as the sublevel set of a Lipschitz-continuous function $l(x)$,
i.e.\ $\mathcal{L} = \{x : l(x) \le 0\}$, most commonly taken to be the signed
distance to the failure set for avoid problems and to the goal set for reach problems.

In this context the Backward Reachable Tube (BRT) refers to the set of states
from which, despite the controller's best effort, the disturbance can force the
system into $\mathcal{L}$ within a time horizon $[t, T]$. The control and
disturbance signals are measurable functions
\begin{equation*}
\begin{aligned}
u(\cdot) &\in \mathcal{U}(t) = \{\phi:[t, T] \rightarrow \mathcal{U} \mid \phi(\cdot) \text{ is measurable} \}, \\
d(\cdot) &\in \mathcal{D}(t) = \{\phi:[t, T] \rightarrow \mathcal{D} \mid \phi(\cdot) \text{ is measurable} \}.
\end{aligned}
\end{equation*}
but they play opposing roles: the controller seeks to keep the system away from
$\mathcal{L}$, while the disturbance acts adversarially to drive it in. To make
this game well posed, the disturbance is modeled as a non-anticipative strategy
$d(\cdot) = \vartheta[u](\cdot)$ with
\begin{equation*}
\begin{aligned}
d \in \Gamma(t) \triangleq \Big\{ \vartheta: \mathcal{U}(t) \rightarrow \mathcal{D}(t) \ \Big| \
&u(r) = \hat{u}(r) \text{ for almost every } r \in [t, s] \\
&\Longrightarrow \vartheta[u](r) = \vartheta[\hat{u}](r) \text{ for almost every } r \in [t, s] \Big\}.
\end{aligned}
\end{equation*}
that is, the disturbance may react to the control applied up to the current
time but cannot anticipate future control inputs. This gives the disturbance an
instantaneous informational advantage, yielding a conservative, worst-case
notion of safety. Throughout, we write $d(\cdot)$ with the understanding that it
implicitly encodes such a non-anticipative response to $u(\cdot)$.

Let $\xi_{x,t}^{u,d}(\tau)$ denote the state at time $\tau \in [t,T]$ of the
trajectory of \eqref{eq:dynamics} starting from $x$ at time $t$ under inputs
$u(\cdot)$ and $d(\cdot)$. %we only use \xi once
The cost function is defined as the minimum of $l$ along the trajectory, 
\begin{equation}
    J\big(x, t, u(\cdot), d(\cdot)\big) = \min_{\tau \in [t, T]} \, l\big(\xi_{x,t}^{u,d}(\tau)\big).
    \label{eq:cost}
\end{equation}

Since $l$ is non-positive exactly on $\mathcal{L}$, taking the minimum over the
horizon records the trajectory's closest approach to the failure set: $J \le 0$
if and only if the trajectory enters $\mathcal{L}$ at some instant in $[t,T]$,
even if it leaves again afterwards. A single violation anywhere along the
trajectory is therefore enough to mark it unsafe.

We are interested in how close to $\mathcal{L}$ the system can be forced under
the best and worst possible behavior: the controller tries to steer the
trajectory as far from the unsafe set as it can manage, while the disturbance
acts adversarially to push it in. Framing this as a zero-sum differential game
gives the value function
\begin{equation}
    V(x, t) = \inf_{d(\cdot)} \sup_{u(\cdot)} \; J(x, t, u(\cdot), d(\cdot)),
    \label{value-function}
\end{equation}
which represents the worst-case margin of safety achievable from $(x,t)$ under
optimal play from both sides.

Applying the dynamic programming principle to \eqref{value-function} shows
that $V$ satisfies a variational form of the Hamilton-Jacobi-Isaacs equation,
known as the HJI-VI:
\begin{subequations}\label{eq:hji-vi}
\begin{align}
    \min\Big\{ D_t V(x,t) + H\big(x,\nabla V(x,t),t\big), \; l(x) - V(x,t) \Big\} &= 0, \label{eq:hji} \\
    V(x, T) &= l(x). \label{eq:terminal}
\end{align}
\end{subequations}

where $D_t V$ and $\nabla V$ are the time and spatial derivatives of the value
function, and the Hamiltonian $H$ is the spatial rate of change of $V$ along the dynamics
under optimal control and worst-case disturbance,
\begin{equation}
    H\big(x,\nabla V(x,t),t\big) = \max_{u} \min_{d} \; \langle \nabla V(x,t), f(x, u, d) \rangle.
    \label{eq:hamiltonian}
\end{equation}

%The $\min\{\cdot\}$ structure of \eqref{eq:hji} is what gives the value functionits "running minimum" character: the second term, $l(x) - V(x,t)$, enforces $V(x,t) \le l(x)$, so a state that becomes unsafe at some point stays marked unsafe for the rest of the horizon, even if the trajectory later moves back toward $\mathcal{L}$'s exterior.

Solving \eqref{eq:hji}-\eqref{eq:terminal} yields the value function, from which
the BRT introduced above is formally recovered as its sub-zero level set,
\begin{equation}
    \mathcal{V}(t) = \{x : V(x,t) \le 0\},
    \label{eq:brt}
\end{equation}
Because $V$ is the worst-case value of $J$ under optimal play, its sign carries
over directly: $V(x,t) \le 0$ means that, even under the controller's best
effort, the disturbance can force the system into $\mathcal{L}$, so $x$ is
unsafe; $V(x,t) > 0$ means the controller can keep the system a strictly
positive distance from $\mathcal{L}$ throughout. The safety-preserving control
policy follows directly from the Hamiltonian's maximizing argument,
\begin{equation}
    u^*(x,t) = \arg\max_{u} \min_{d} \; \langle \nabla V(x,t), f(x,u,d) \rangle.
    \label{eq:optimal-control}
\end{equation}
As long as the system which starts outside the BRT applies the control in \eqref{eq:optimal-control} at the BRT boundary, the system maintains safety.

In practice, solving the HJI-VI requires discretizing the state space and time
horizon into a grid, and the cost of doing so grows exponentially with the state
dimension, this is what confines classical HJ reachability to lower dimensional systems.
\subsubsection{DeepReach}
DeepReach \cite{deepreach} sidesteps the grid by representing the value
function with a neural network $V_\theta(x,t)$ that takes the state and time as
input. The network is trained in a self-supervised manner, by minimizing the
residual of the HJI-VI \eqref{eq:hji-vi} at randomly sampled state--time pairs,
so no ground-truth value function is required. Because it never discretizes
the state space, DeepReach avoids the exponential cost of grid-based solvers
\cite{mitchell2004toolbox, hj_reach_ASL2023}: its memory and training cost
depend mainly on the complexity of the value function rather than on the state
dimension, which makes systems well beyond the reach of grids tractable. 
%parameter count vs grid points is a bit ornages vs apples comparison, i pivoted towards compute time which is the common ground

There has been recent updates to the DeepReach based framework, \cite{exact} imposes the safety boundary conditions exactly during the learning process by restructuring the Value function as the combination of target function and a neural network output. Building on this, \cite{hopf} leverages rapidly computed linear PDE solutions to guide the neural training process. Model Predictive Control(MPC) has been used to warm start the learning process for the neural network and by iteratively refining the MPC approximations, the reachability solution is learned which is formulated as a single player game \cite{MPCD} and it has been extended to the two player zero sum game in MPC guided Adversarial DeepReach~\cite{MPCAD}. Gradient free methods based on exploiting the bang-bang structure of optimal control and disturbances has also been proposed to train neural networks to approximate HJ Reachability solutions \cite{gradfree}.
\subsection{State Estimation Methods (Continuous Time)}
\label{sec:ct_extended_estimators}
% \noindent\textit{Notation here is generic; it is specialized in Section~\ref{sec:psetup}.} \\
Consider a nonlinear system with known input \(u(t)\), state \(X(t)\), measurement \(Y(t)\), and exogenous process and measurement disturbances \(w(t)\) and \(n(t)\):
\begin{align}
    \dot{X}(t) &= g(X(t),u(t),t) + B(t)w(t), \\
    Y(t) &= h(X(t),u(t),t) + D(t)n(t), \\
    Z(t) &= m(X(t)),
\end{align}
where \(Z(t)=m(X(t))\) is a nonlinear function of the state that is desired to be estimated. The vector-valued functions \(g(\cdot)\), \(h(\cdot)\), and \(m(\cdot)\) are assumed differentiable with continuous partial derivatives.

The extended estimator is obtained by linearizing the dynamics and measurement equations around the current estimate. Throughout, a hat denotes an estimated
quantity: \(\hat{X}(t)\) is the estimate of \(X(t)\) computed from the inputs
and measurements. Define
\begin{equation}
    A(t) = \left.\frac{\partial g}{\partial X}\right|_{X=\hat{X}(t)}, \qquad
    C(t) = \left.\frac{\partial h}{\partial X}\right|_{X=\hat{X}(t)}, \qquad
    L_m(t) = \left.\frac{\partial m}{\partial X}\right|_{X=\hat{X}(t)}.
\end{equation}

\subsubsection{Continuous-Time Extended Kalman Filter (CT-EKF)}
In the CT-EKF, \(w(t)\) and \(n(t)\) are zero-mean white noises with covariances \(Q\) and \(R\), and the filter minimizes the expected estimation error. We therefore identify the disturbance channels of the base system with these noises by taking
\begin{equation}
    B(t)B(t)^\top = Q, \qquad D(t)D(t)^\top = R,
    \label{eq:ekf-noise-identification}
\end{equation}
so that \(B(t)\) and \(D(t)\) play the role of the process and measurement-noise weighting matrices. The output map is the identity, \(m(X)=X\), so that \(Z(t)=X(t)\), \(\hat{Z}(t)=\hat{X}(t)\), and \(L_m(t)=\partial m/\partial X=I\). 
With \eqref{eq:ekf-noise-identification}, the CT-EKF equations~\cite{DanSimon} are:
\begin{align}
    \dot{\hat{X}} &= g(\hat{X},u,t) + K\big[Y-h(\hat{X},u,t)\big], \label{eq:ctekf-mean} \\
    K &=   \Sigma C^\top  R^{-1}, \label{eq:ctekf-gain} \\
    \dot{ \Sigma} &= A \Sigma+\Sigma A^\top+Q- \Sigma C^\top R^{-1}C \Sigma, \label{eq:ctekf-cov}
\end{align}
with \( \Sigma(0)=\mathbb{E}\big[(X(0)-\hat{X}(0))(X(0)-\hat{X}(0))^\top\big]\). Here \(\Sigma(t)\) is the state-estimation error covariance obeying the standard continuous-time Riccati covariance update \eqref{eq:ctekf-cov}.
\subsubsection{Continuous-Time Extended H$_\infty$ Filter}
The H$_\infty$ filter minimizes the worst-case estimation error against unknown disturbances, rather than the expected error under assumed noise statistics as in the Kalman filter \cite{H_infinity}. The continuous-time extended H$_\infty$ filter \cite{ehinf} has the same structure as the CT-EKF, but treats $w(t)$ and $n(t)$ as unknown bounded disturbances, so the matrices $BB^\top = Q$ and $DD^\top = R$ of \eqref{eq:ekf-noise-identification} act as weights in the Riccati equation rather than as noise covariances. With the same identity output map, $m(X) = X$ and $L_m = I$, the filter equations are

\begin{align}
    \dot{\hat{X}} &= g(\hat{X},u,t) + K\big[Y-h(\hat{X},u,t)\big], \label{eq:hinf-mean} \\
    K &=  P C^\top \big(D D^\top\big)^{-1}, \label{eq:hinf-gain} \\
    \dot{P} &= A P+ P A^\top -  P C^\top\big(D D^\top\big)^{-1}C P + \frac{1}{\gamma^2} P L_m^\top L_m  P + BB^\top. \label{eq:hinf-cov}
\end{align}
The Riccati equation \eqref{eq:hinf-cov} must be integrated together with the state-estimate equation at runtime to generate \(\hat{Z}(t)\). In this case $P$ is not the error covariance matrix, but it provides an upper bound on it \cite{upperbound}; we use $P$ in place of $\Sigma$ wherever the $H_\infty$ filter is concerned.

% \noindent
% Comparing \eqref{eq:ctekf-mean}-\eqref{eq:ctekf-cov} and \eqref{eq:hinf-mean}-\eqref{eq:hinf-cov}, the CT-EKF and the CT extended \(H_\infty\) filter share the same extended structure and gain form. The CT-EKF is recovered as the stochastic special case with the attenuation term dropped (equivalently \(\gamma\to\infty\)), whereas the \(H_\infty\) filter retains a finite \(\gamma\) and the additional robust term \(\tfrac{1}{\gamma^2} P L_m^\top L_m  P\).
\section{Problem Formulation}
\label{sec:psetup}

\paragraph{System and measurements.}
We consider a system with true state $\mathbf{x}=(x,x_u)\in\mathbb{R}^n$, control
$u\in\mathcal{U}$, and bounded physical disturbance $d_{\mathrm{phys}}\in\mathcal{D}$,
\begin{equation}
    \dot{\mathbf{x}} = f(\mathbf{x},u,d_{\mathrm{phys}}),
\end{equation}
where $x$ is perfectly known and $x_u\in\mathbb{R}^{n_{\hat{x}}}$ is observed only through
\begin{equation}
    y = h(x_u)+n, \qquad n\sim\mathcal{N}(0,R).
\end{equation}

\paragraph{Given estimator.}
We assume a state estimator is provided and \emph{fixed}. It outputs an estimate $\hat{x}$ of $x_u$ and an error covariance
$\Sigma$, so that $x_u\sim\mathcal{N}(\hat{x},\Sigma)$. The controller has access only
to $(x,\hat{x},\Sigma)$, never to $x_u$. The
estimate is driven by the measurement innovation $e = y - h(\hat{x})$:
\begin{equation}
    \dot{\hat{x}} = g(\hat{x}, u, e,  d_{\mathrm{phys}} ), \qquad
    \dot{\Sigma} = q(\hat{x}, \Sigma, u),
\end{equation}
where $q$ is the Riccati covariance update with process-noise covariance $Q$.

\paragraph{Safety specification.}
Let $\mathcal{L}_{\mathbf{x}}\subset\mathbb{R}^n$ be the set of true failure states (e.g., collision).
The system is safe over $[t,T]$ if $\mathbf{x}(\tau)\notin\mathcal{L}_{\mathbf{x}}$ for all
$\tau\in[t,T]$. Since $x_u$ is unobserved, safety must be certified from
$(x,\hat{x},\Sigma)$ alone.

\begin{figure}[t]
\captionsetup{font=footnotesize}
    \centering
    \includegraphics[width=0.8\linewidth]{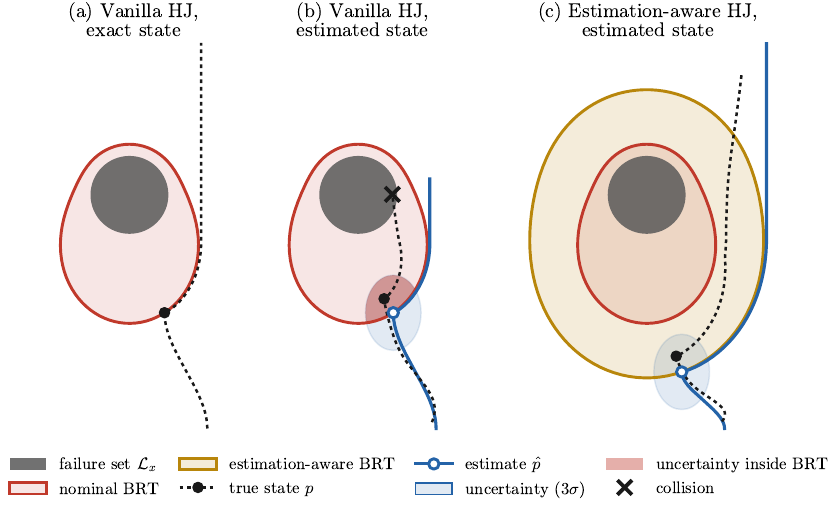}
    \caption{Safety filtering with (a) vanilla HJ and the exact state, (b) vanilla
HJ and an estimate $\hat p$, and (c) estimation-aware HJ under the same
estimation error as (b).}
    \label{fig:motivation}
\end{figure}

\paragraph{Why vanilla HJ reachability fails.}
Standard HJ reachability computes the unsafe set $\mathcal{V}_{\mathbf{x}}$ of the failure set $\mathcal{L}_{\mathbf{x}}$ assuming the full state is known~\cite{hjr}, and a filter engaging on its boundary keeps the system safe (Fig.~\ref{fig:motivation}a).
Under partial observability, a naive approach is to apply the same filter to the estimate, which is unsound for two reasons
(Fig.~\ref{fig:motivation}b). First, when the filter engages, the true state may already be inside $\mathcal{V}_{\mathbf{x}}$, a risk that grows with sensor noise.
Second, the estimation error keeps changing as the estimator reacts to new measurements and disturbances, so a margin that covers it at one moment may not cover it over the whole horizon.
Inflating $\mathcal{V}_{\mathbf{x}}$ by a fixed margin would compensate only at the cost of needless conservatism where the estimate is accurate.
We instead account for the estimator's uncertainty and its evolution within the reachability problem, which engages the filter while likely true states are still safe (Fig.~\ref{fig:motivation}c).

% \begin{figure}[t]
%     \centering
%     \includegraphics[width=0.8\linewidth]{figures/motivation1.png}
%     \caption{ Left (Standard HJR): Vanilla HJ reachability treats the estimate as the true state.
%     The planned trajectory ends on the BRT boundary which is on the edge of unsafe boundary.
%     Right (Estimated states setting): The same trajectory with the estimator's uncertainty ellipse
%     ($\Sigma$) overlaid. The actual value of the state could be anywhere in the uncertainty ellipse(region), and hence the state could either be in the safe or the unsafe region already which emphasizes the need for an estimation aware HJR }
%     \label{fig:motivation}
% \end{figure}

\paragraph{Problem statement.}
%might need to change after guarantees section
Given $(f,h,\mathcal{L}_{\mathbf{x}})$, the fixed estimator, a horizon $[t,T]$, and a confidence
level $\delta\in(0,1)$, find a set $\mathcal{S}$ of information states $(x,\hat{x},\Sigma)$
and a safe control strategy such that, whenever the initial information state lies in
$\mathcal{S}^c$, the information state is kept in $\mathcal{S}^c$ over $[t,T]$ for all
admissible $d_{\mathrm{phys}}\in\mathcal{D}$ and all admissible innovations, and, at every
instant $\tau\in[t,T]$ at which the information state lies in $\mathcal{S}^c$, the true
state satisfies $\mathbf{x}(\tau)\notin\mathcal{L}_{\mathbf{x}}$ with probability at least $1-\delta$
(a \emph{pointwise-in-time} guarantee). A trajectory-level guarantee, i.e.\ avoiding
$\mathcal{L}_{\mathbf{x}}$ over the whole interval $[t,T]$ with probability at least $1-\delta$, is
the stronger goal; we do not establish it here and discuss it in
Section~\ref{sec:conclusion}.
We want $\mathcal{S}$ to satisfy:
\begin{enumerate}
    \item \textbf{Uncertainty-aware:} $\mathcal{S}$ grows with $\Sigma$, adding conservativeness to
          states where the estimator is unsure.
    \item \textbf{Consistent with vanilla HJ reachability:} $\mathcal{S}\to\mathcal{V}_{\mathbf{x}}$ when the state
          becomes perfectly known, i.e.\ when both the initial uncertainty and the process
          noise vanish ($\Sigma(t)\to0$ and $Q\to0$).
          % see the Remark in
          % Section~\ref{sec:method}. Vanishing initial uncertainty alone is not sufficient, since
          % $Q\neq0$ makes $\Sigma$ grow over the horizon (Section~\ref{sec:air3d}).
    \item \textbf{Estimator-aware:} accounts for how the fixed estimator's error and
          covariance evolve over the horizon, not only their current values.
    % \item \textbf{Computable:} Computed offline for $n_{\hat{x}}$ of practical
    %       size (we use neural approximations of the value function\cite{deepreach, MPCAD}).
\end{enumerate}

\section{Estimation Aware Reachable Sets}
\label{sec:method}
With the division of states into perfectly known and unknown states, the controller therefore acts on the belief state
\begin{equation}
    z = \begin{bmatrix} x \\ \hat{x} \\ \operatorname{vec}(\Sigma) \end{bmatrix}
    \in \mathcal{Z}
\end{equation}
rather than on the true state. Here, $\operatorname{vec}(\cdot)$ denotes the half-vectorization, which stacks the $n(n+1)/2$ unique entries of a symmetric matrix (its upper-triangular part, column by column) into a vector, i.e., $\operatorname{vec}:\mathbb{S}^{n}_{+}\to\mathbb{R}^{n(n+1)/2}$.

\paragraph{Augmented dynamics:} The belief state evolves as
\begin{equation}
    \dot z = F(z, u, d_{\mathrm{phys}}, e) =
    \begin{bmatrix}
        f\big(x, u, d_{\mathrm{phys}}\big) \\
        g(\hat{x}, u, e, d_{\mathrm{phys}}) \\
        \operatorname{vec}\big(q(\hat{x}, \Sigma, u)\big)
    \end{bmatrix}.
    \label{eq:augdyn}
\end{equation}
The disturbance $d_{\mathrm{phys}}$ is constrained to $\mathcal{D}$, while $e$
is constrained to the admissible innovation set $\mathcal{E}(\Sigma)$, which is
therefore state-dependent through $z$. 
%The choice of $\mathcal{E}(\Sigma)$ is specified in the next section.

\paragraph{Failure set:} Let $\mathcal{L}_{x} \subset \mathbb{R}^n$ denote the
set of true failure states (e.g., collision). Because $x_u$ is never known
exactly, safety cannot be evaluated from $\hat{x}$ alone: the failure set in
terms of the estimate must depend on how uncertain that estimate is. We
therefore seek a failure set $\mathcal{L}_z \subseteq \mathcal{Z}$ over the
belief state $z$, with $\mathcal{L}_z = \{z : \ell(z) \le 0\}$.

\paragraph{Objective:} Standard HJ reachability computes the BRT of $\mathcal{L}_{x}$ assuming the full state is known. We instead
seek the set of belief states from which the system may reach the target set
$\mathcal{L}_z$ within the horizon $[t,T]$ under the estimation uncertainty.
Let $\zeta(\tau;\, z, t, u(\cdot), d(\cdot))$ denote the trajectory of the
augmented dynamics at time $\tau$, starting from $z$ at time $t$ under control
$u(\cdot)$ and adversary $d(\cdot) = (d_{\mathrm{phys}}(\cdot), e(\cdot))$,
with $d_{\mathrm{phys}} \in \mathcal{D}$ and $e \in \mathcal{E}(\Sigma)$. The BRT of $\mathcal{L}_z$ is
\begin{equation}
\resizebox{0.9\linewidth}{!}{$
\mathcal{V}(t) = \left\{ z \in \mathcal{Z} :
\exists\, \vartheta \in \Gamma(t),\;
\forall\, u(\cdot) \in \mathcal{U}(t),\;
\exists\, \tau \in [t,T],\;
\zeta\big(\tau;\, z, t, u(\cdot), \vartheta[u](\cdot)\big) \in \mathcal{L}_z \right\}
$},
\end{equation}
where, as in Section~\ref{sec:background}, $\Gamma(t)$ is the set of non-anticipative
strategies of the adversary, now mapping $u(\cdot)$ to signals
$d(\cdot) = (d_{\mathrm{phys}}(\cdot), e(\cdot))$ with $d_{\mathrm{phys}}(\tau)\in\mathcal{D}$ and
$e(\tau)\in\mathcal{E}(\Sigma(\tau))$ for almost every $\tau$. The adversary may thus react to
the control applied so far but not to future control inputs.
For the avoid case, $\mathcal{V}(t)$ is the set of belief states that should be
avoided and vice versa for the reach case. 

We propose an HJ reachability formulation with a covariance-dependent bound.
The estimation error $x_u - \hat{x}$ is never available to the controller,
since it would require knowing $x_u$. The closest observable proxy is the
measurement innovation, which drives the estimator update. As declared in
Section~\ref{sec:psetup}, we treat $e$ as an adversarial variable and, to turn the estimator update into an adversarial game for HJ reachability,
we restrict it to a set determined by the estimator state. With measurement matrix
$C$ and measurement noise covariance $R$, the innovation
covariance is $S = C\Sigma C^\top + R$. Motivated by this, we define  the admissible innovation set as the box
\begin{equation}
    \mathcal{E}(\Sigma) = \left\{ e \in \mathbb{R}^{m} \;\middle|\; |e_i| \le \kappa \sqrt{S_{ii}}, \quad S = C\Sigma C^\top + R, \quad i = 1, \dots, k \right\},
\end{equation}
where $k$ is the measurement dimension.\\
The innovation is a white-noise process, which is
unbounded and therefore cannot be handled directly by a worst-case formulation.
We instead treat it as an adversarial input confined to $\mathcal{E}(\Sigma)$.
This is a deliberate conservative modeling choice: the adversary may select the
most harmful admissible innovation at every instant, including realizations that
a stochastic innovation would rarely produce, so the resulting reachable set
over-approximates the effect of the correction term and errs on the side of
safety. The parameter $\kappa$ controls the degree of conservatism.
The HJI-VI equation~\cite{mitchell2005time} is formed as:
\begin{equation}
    \frac{\partial V}{\partial t} + \max_{u \in \mathcal{U}} \min_{d \in \mathcal{D},\, e \in \mathcal{E}(\Sigma)} \Big( \nabla_z V \cdot F(z, u, d, e) \Big) = 0,
    \qquad V(z, T) = \ell(z).
\end{equation}

Furthermore, the failure set must be modified to incorporate the state uncertainty, ensuring that the BRT accounts for the estimation uncertainty as it evolves backward in time. Concretely, we replace the nominal target set $\ell(x)$ with an inflated failure set $\ell(z)$ defined over the augmented state,  such that the boundary of the unsafe set adapts to the magnitude and structure of the estimation uncertainty. 
%The exact form of this modification is presented in the context of Air3D formulation.

\noindent\textbf{Target/failure set:} We inflate the nominal collision set by an uncertainty margin scaled by the confidence parameter $\kappa$.
% (chosen as $\kappa = 2$, corresponding to roughly 95\% coverage under a Gaussian belief)
 % Two variants are considered.

Let $p$ denote the position components of the estimated state $x_u$, $\hat{p}$ their estimate, and
$\Sigma_{pp}$ the corresponding $n_p\times n_p$ block of $\Sigma$. Given the nominal
implicit failure function $l(x,p)$, which is $1$-Lipschitz in $p$, we define the
estimation-aware target by using $\lambda_{\max}$ the largest (worst-case) eigenvalue of
$\Sigma_{pp}$:
\begin{equation}
\label{eq:generic-target}
\ell(z) = l(x,\hat{p}) - \kappa\sqrt{\lambda_{\max}(\Sigma_{pp})}.
\end{equation}
For avoid problems this inflates the failure set by $\kappa\sqrt{\lambda_{\max}(\Sigma_{pp})}$;
for reach problems the same margin with the opposite sign erodes the target.
Using this formulation on a HJ reachability solving pipeline, we obtain both the Value function and its gradients, which can then be used to implement a state estimation aware safety filter that adapts its behavior based on the observed variance of the estimation error.

\noindent\textbf{Remark:} Consider the noise-free process case ($Q=0$, $\Sigma(0)=0$).
The Riccati dynamics give $\dot\Sigma|_{\Sigma=0}=Q=0$, so $\Sigma(t)\equiv 0$, and
the target margin $\kappa\sqrt{\lambda_{\max}(\Sigma)}$ vanishes, giving
$\ell(z)=\ell(\hat{x})$. The set $\mathcal{E}(0)$ is not $\{0\}$
because $R\neq 0$, but the filter gain is $K=\Sigma C^\top R^{-1}=0$, so the innovation
has no effect on $\dot{\hat{x}}$. The Hamiltonian therefore reduces to
$\max_u\min_d \nabla_x V\cdot f(\hat{x},u,d)$, and the formulation reduces to the
standard HJI-VI on the slice $\Sigma=0$.

\noindent\textbf{Probabilistic Safety Guarantee:}
The formulation above is a robust game: the innovation $e$ is confined to
$\mathcal{E}(\Sigma)$ by construction, and the value function $V$ of
\eqref{eq:hji-vi} certifies safety in belief space for that game. We now relate
this certificate to the true state. Let $\tilde{x}(t) := x_u(t) - \hat{x}(t)$
denote the estimation error, which the controller does not observe, and let
$\tilde{p}$ denote its position components, so that $p = \hat{p} + \tilde{p}$.
The failure set is described by a failure margin
$\ell_x:\mathbb{R}^{n_p}\to\mathbb{R}$ on the position,
$\mathcal{L}_x = \{x : \ell_x(p) \le 0\}$, which we assume to be
$L_\ell$-Lipschitz. A positive value
certifies only the belief: it says nothing deterministic about the true
position, because $\tilde{p}$ is unknown. The guarantee on the true state is
therefore probabilistic, with the probability coming from the estimation error
$\tilde{p}$, under the assumption that the estimator is consistent, i.e.,
$\Sigma$ is the true error covariance.\\

\noindent\textbf{Pointwise probabilistic non-collision guarantee:}
\begin{theorem}
\label{thm:pointwise}
Assume $\Sigma_{pp}(t) \succ 0$ and that the position estimation error is
Gaussian given the belief,
$\tilde{p}(t) \mid z(t) \sim \mathcal{N}\big(0,\Sigma_{pp}(t)\big)$. Fix
$t \in [t_0,T]$ and let the target set be the uncertainty-inflated failure set
\begin{equation}
\label{eq:circular}
\ell(z) = \ell_x(\hat{p}) - L_\ell\,\kappa\sqrt{\lambda_{\max}(\Sigma_{pp})}.
\end{equation}
For the circular failure set,$ L_\ell=1$ in \eqref{eq:generic-target}\\
If $V(z(t),t) > 0$, then
\begin{equation}
    \mathbb{P}\big(x_u(t) \notin \mathcal{L}_x \,\big|\, z(t)\big)
    \;\ge\; 1 - \delta_x(\kappa),
    \qquad
    \delta_x(\kappa) := \mathbb{P}\big(\chi^2_{n_p} > \kappa^2\big),
    \label{eq:pointwise}
\end{equation}
where $n_p = \dim(p)$. The bound is pointwise in time, involves no
horizon or union bound, and does not depend on the innovation.
\end{theorem}

\begin{proof}
Define the events
\begin{equation}
    \Omega_\kappa := \big\{\|\Sigma_{pp}(t)^{-1/2}\tilde{p}(t)\| \le \kappa\big\},
    \qquad
    \Omega_{\mathrm{safe}} := \big\{x_u(t) \notin \mathcal{L}_x\big\},
\end{equation}
i.e., the estimation error lies in the $\kappa$-confidence ellipsoid, and the
true state is outside the failure set. Since $V \le \ell$ in the avoid case,
$V(z(t),t) > 0$ implies $\ell(z(t)) > 0$, so by \eqref{eq:circular}
\begin{equation}
    \ell_x(\hat{p}(t)) > L_\ell\,\kappa\sqrt{\lambda_{\max}(\Sigma_{pp}(t))}.
\end{equation}
On $\Omega_\kappa$,
\begin{equation}
    \|\tilde{p}(t)\|
    \le \sqrt{\lambda_{\max}(\Sigma_{pp}(t))}\,\|\Sigma_{pp}(t)^{-1/2}\tilde{p}(t)\|
    \le \kappa\sqrt{\lambda_{\max}(\Sigma_{pp}(t))},
\end{equation}
and therefore, by Lipschitz continuity of $\ell_x$,
\begin{equation}
    \ell_x(p(t)) = \ell_x\big(\hat{p}(t) + \tilde{p}(t)\big)
    \ge \ell_x(\hat{p}(t)) - L_\ell\,\|\tilde{p}(t)\| > 0,
\end{equation}
so $\Omega_\kappa \subseteq \Omega_{\mathrm{safe}}$. Since $V(z(t),t)$ is a
deterministic function of $z(t)$, this inclusion holds under the conditional
measure given $z(t)$, and hence
$\mathbb{P}(\Omega_{\mathrm{safe}} \mid z(t)) \ge \mathbb{P}(\Omega_\kappa \mid z(t))$.

Under the Gaussian assumption, the whitened error
$w(t) := \Sigma_{pp}(t)^{-1/2}\tilde{p}(t)$ satisfies
$w(t) \mid z(t) \sim \mathcal{N}(0, I)$, so
$\|w(t)\|^2 \mid z(t) \sim \chi^2_{n_p}$ and
\begin{equation}
    \mathbb{P}(\Omega_\kappa \mid z(t)) = \mathbb{P}\big(\chi^2_{n_p} \le \kappa^2\big)
    = 1 - \delta_x(\kappa).
\end{equation}
Combining the two gives \eqref{eq:pointwise}.
\end{proof}

\noindent\textbf{Remark (role of the value function):}
Theorem~\ref{thm:pointwise} uses $V$ only through $V \le \ell$: its guarantee
holds at any time at which the belief lies outside the inflated target set
\eqref{eq:circular}. The value function is what keeps the belief there. 
If $V(z(t_0),t_0) > 0$ and the safety-preserving controller is applied, $V$
remains positive, and hence \eqref{eq:pointwise} holds at every
$t \in [t_0,T]$. This is because the innovation is treated adversarially within
$\mathcal{E}(\Sigma)$: the worst-case choice pushes the unsafe boundary further
out than a random innovation would. The resulting
set is therefore conservative, with $\kappa$ controlling the degree.

\noindent\textbf{Remark (covariance upper bounds):}
If the exact error covariance is unavailable and a known upper bound
$P_{pp}(t)\succeq\Sigma_{pp}(t)$ is used instead to construct the target set
\eqref{eq:circular}, the same guarantee holds with
$\Omega_\kappa := \{\|P_{pp}(t)^{-1/2}\tilde{p}(t)\|\le\kappa\}$: the inclusion
$\Omega_\kappa \subseteq \Omega_{\mathrm{safe}}$ is unchanged, and Anderson's
inequality~\cite{anderson} for this symmetric convex set replaces the exact
$\chi^2_{n_p}$ step, so the right-hand side of \eqref{eq:pointwise} remains a
valid lower bound. For the $H_\infty$ filter, $P_{pp}(t)$ upper-bounds the
error covariance of the linearized filter, so the theorem applies to the
extended filter to the extent that this bound holds along the trajectory.

% If
% $V(z(t_0),t_0) > 0$ and the safety-preserving controller is applied, $V$
% remains positive, and hence \eqref{eq:pointwise} holds at every
% $t \in [t_0,T]$, as long as the realized innovation stays in
% $\mathcal{E}(\Sigma)$. Innovations outside $\mathcal{E}(\Sigma)$ can drive $V$
% to non-positive values, in which case the premise of the theorem fails.

% The innovation, however, is a zero-mean Gaussian white-noise process. The formulation treats it as an adversary restricted to $\mathcal{E}(\Sigma)$, which is conservative: the adversary may select the most harmful admissible innovation at every instant, including sustained worst-case realizations that a Gaussian innovation would rarely produce. Within the box this over-approximates the effect of the correction term, so the guarantee on the true state $x_u$ holds with high probability rather than surely, under the assumption that the estimator is consistent, i.e., $\Sigma$ is the true error covariance.
\section{Case Studies}
\label{sec:case}
\subsection{Air3D with CT-EKF}
\label{sec:air3d}
\noindent\textbf{Setup:} We consider the pursuer-evader collision avoidance problem \cite{hjr}, in which the relative heading $\psi$ is known and the pursuer's position $(\hat{x},\hat{y})$ relative to the evader, expressed in the evader's body frame, is tracked by a CT-EKF. The augmented state is $z = [\hat{x}, \hat{y}, \psi, \Sigma_{xx}, \Sigma_{yy}, \Sigma_{xy}]^\top$ with
\begin{align}
\dot{\hat{x}} &= v_P\cos\psi - v_E + u_E\hat{y} + K_{11}e_1 + K_{12}e_2, \\
\dot{\hat{y}} &= v_P\sin\psi - u_E\hat{x} + K_{21}e_1 + K_{22}e_2, \\
\dot{\psi} &= u_P - u_E,
\end{align}
where $e$ is the innovation and $K=\Sigma/r$. The covariance follows the CT-EKF Riccati equation \eqref{eq:ctekf-cov} with $A = \begin{bmatrix} 0 & u_E \\ -u_E & 0 \end{bmatrix}$, $C = I_2$, $R = rI_2$, $Q = qI_2$ (written out in \eqref{eq:covariance}, Appendix~\ref{app:air3d}). The adversary is $(u_P, e)$ with $e\in\mathcal{E}(\Sigma)$.

\noindent\textbf{Parameters:} Both agents move at $v_P = v_E = 0.75$ with turn rates $|u_E| \le 4$ and $|u_P| \le 3$, the collision set is a disc of radius $R_c = 0.25$, and $q = r = 0.05$, intentionally large so that the effect of the estimator is visible. The target uses the circular bound \eqref{eq:circular} with $\kappa = 3$, so with $n_p=2$ Theorem~\ref{thm:pointwise} gives $\mathbb{P}(x_u(t) \notin \mathcal{L}_x \mid z(t)) \ge 1 - e^{-\kappa^2/2} \approx 0.989$ whenever $V(z(t),t)>0$. The value function is trained with DeepReach and MPC supervision \cite{MPCAD} over the horizon $t\in[0,1]$\,s on the domain $\hat x,\hat y\in[-3,3]$, $\psi\in[-\pi,\pi]$, $\Sigma_{xx},\Sigma_{yy}\in[0.001,0.05]$ and $\Sigma_{xy}\in[-0.025,0.025]$; the upper covariance bound is the steady state $\sqrt{qr}=0.05$ of the isotropic Riccati equation, and the training settings are in Table~\ref{tab:hyper}.

\subsubsection{Results}

\begin{figure}[t]
  \captionsetup{font=footnotesize}
  \centering
  \begin{minipage}[c]{0.35\linewidth}
    \centering
    \includegraphics[width=\linewidth]{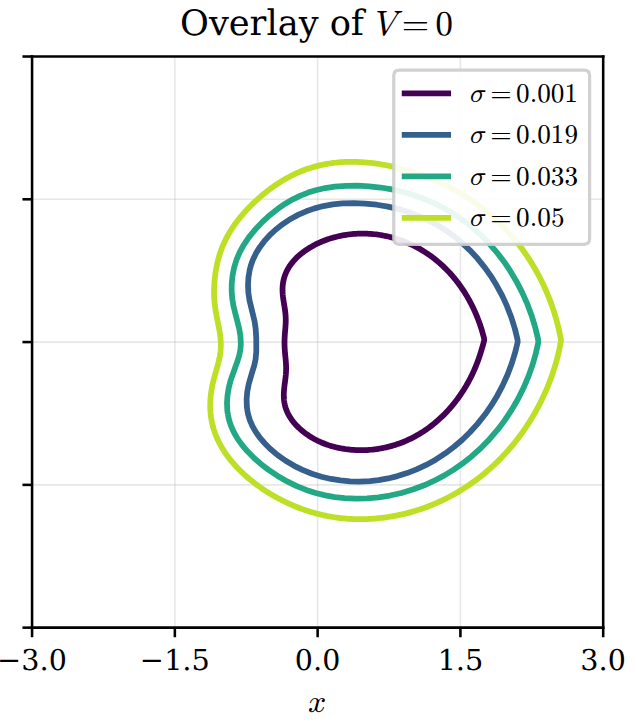}
    \caption{Learned estimation-aware BRT at $t = T$ and $\psi = \pi$ for
    $\sigma \in \{0.001, 0.019, 0.033, 0.05\}$ with $\Sigma_{xy} = 0$.}
    \label{fig:brt_sigma}
  \end{minipage}\hfill
  \begin{minipage}[c]{0.5\linewidth}
    \centering
    \captionof{table}{Air3D safety filter against the worst-case pursuer ($7500$ rollouts per row, $5$\,s horizon). Coll.: true distance below $R_c$. Interv.: rollouts in which the filter engaged; first intervention $t$ is averaged over those rollouts.}
    \label{tab:rollout_stats_offset03}
    \footnotesize
    \setlength{\tabcolsep}{3pt}
    \resizebox{\linewidth}{!}{%
    \begin{tabular}{llccc}
      \toprule
      Filter & $\Sigma_0$ & \makecell{Coll.\\(\%)} & \makecell{Interv.\\(\%)} & \makecell{First\\interv. (s)} \\
      \midrule
      \multirow{3}{*}{No filter} & 0.001 & 31.4 & -- & -- \\
       & 0.025 & 28.0 & -- & -- \\
       & 0.05  & 25.4 & -- & -- \\
      \midrule
      \multirow{3}{*}{Nominal} & 0.001 & 30.4 & 22.8 & 1.38 \\
       & 0.025 & 26.9 & 21.1 & 1.61 \\
       & 0.05  & 24.3 & 19.7 & 1.81 \\
      \midrule
      \multirow{3}{*}{\makecell[l]{Nominal +\\margin (ss)}} & 0.001 & 10.7 & 76.8 & 0.74 \\
       & 0.025 & 8.5 & 55.9 & 1.51 \\
       & 0.05  & 7.1 & 46.7 & 1.74 \\
      \midrule
      \multirow{3}{*}{\makecell[l]{Nominal +\\margin (max)}} & 0.001 & 7.2 & 85.5 & 0.59 \\
       & 0.025 & 5.0 & 64.3 & 1.32 \\
       & 0.05  & 4.1 & 52.7 & 1.63 \\
      \midrule
      \multirow{3}{*}{\makecell[l]{Estimation-\\aware}} & 0.001 & \textbf{2.5} & 100.0 & 0.16 \\
       & 0.025 & \textbf{0.3} & 98.8 & 0.38 \\
       & 0.05  & \textbf{0.1} & 91.3 & 0.63 \\
      \midrule
      \multirow{3}{*}{Belief CBF} & 0.001 & 7.7 & 76.1 & 0.81 \\
       & 0.025 & 5.2 & 60.6 & 0.79 \\
       & 0.05  & 3.8 & 51.4 & 0.93 \\
      \bottomrule
    \end{tabular}%
    }
  \end{minipage}
\end{figure}

\noindent\textbf{BRT growth with estimation uncertainty:} Fig.~\ref{fig:brt_sigma} shows $(\hat x,\hat y)$ slices of the learned BRT $\{V\le0\}$ at $\psi=\pi$ and isotropic covariance $\Sigma_{xx}=\Sigma_{yy}=\sigma$, $\Sigma_{xy}=0$. The BRT grows with $\sigma$, and not as a uniform inflation of the nominal set: it reflects the joint evolution of the state and the estimation error. On this slice the 6D problem reduces exactly to a 4D one, which we solve on a grid with \texttt{hj\_reachability} \cite{hj_reach_ASL2023}; the learned and grid BRTs agree with IoU $0.92$--$0.95$ (Appendix~\ref{app:air3d_grid}). The estimation-aware BRT contains the nominal Air3D BRT, computed on a grid with the same toolbox, on every slice we evaluated, and its $(\hat x,\hat y)$ area is on average $13.6\times$ larger at $\sigma=0.001$ and $37\times$ larger at $\sigma=0.05$.

% \begin{wrapfigure}{r}{0.35\linewidth}
%   \centering
%   \vspace{-1em}
%   \includegraphics[width=\linewidth]{figures/pi_brt.png}
%   \caption{Learned estimation-aware BRT at $t = T$ and $\psi = \pi$ for
%   $\sigma \in \{0.001, 0.019, 0.033, 0.05\}$ with $\Sigma_{xy} = 0$.
%   Left: filled sets, where red marks the inside of the BRT ($V \le 0$).
%   Right: overlay of the $V = 0$ boundaries.}
%   \label{fig:brt_sigma}
%   \vspace{-1em}
% \end{wrapfigure}

\noindent\textbf{Safety filter:} The evader holds a north heading, and a least-restrictive filter~\cite{hjrsafety,Fisac2024safe_filter}. The pursuer observes the true state and plays the worst-case turn rate of the nominal Air3D game. We compare no filter, the nominal Air3D BRT, the nominal BRT with a constant margin $\kappa\sqrt{\bar\sigma}$ ($\bar\sigma=0.05$, steady state, or $0.075$, maximum), the estimation-aware BRT, and belief CBF \cite{bcbf}. For each $\Sigma_0=\sigma I_2$, $\sigma\in\{0.001,0.025,0.05\}$, we run $7500$ rollouts ($15$ initial positions $0.3$\,m outside the estimation-aware BRT, $500$ noise seeds each), shared by all filters; details are in Appendix~\ref{app:air3d}.

% preamble

\noindent\textbf{Results:} In Table~\ref{tab:rollout_stats_offset03}, the nominal filter lowers collisions by only about one percentage point because it engages late when the pursuer is too close ($1.4$--$1.8$\,s). The estimation-aware filter engages at $0.16$--$0.63$\,s and drastically cuts collision rates from $(31.4\%, 28.0\%, 25.4\%)$ to $($2.5\%, 0.3\%, $0.1\%)$, whereas the constant-margin and belief-CBF baselines, which do not anticipate how the covariance and innovation evolve, collide with rates between $3.8$--$10.7\%$. The constant-margin filters are computed from the nominal game, which does not model the innovation. During the encounter, estimator corrections push the
estimate into the inflated set even under the optimal safe control, so the
margin is gradually used up causing higher collision rates. The estimation-aware filter avoids this at the
cost of conservatism, in this example it engages in $91$--$100\%$ of rollouts, a consequence of
the intentionally large sensor noise and the worst-case treatment of the
innovation.

\subsection{Planar Quadrotor with $H_\infty$ filter}
\label{sec:quad}

\textbf{Setup:} We consider a planar quadrotor with state $[x,y,\theta,v_x,v_y,\omega]^\top$ and thrust inputs $u_1,u_2$, where the attitude $(\theta,\omega)$ and velocity $v$ are known and only the position $p=[x,y]^\top$ is estimated. Since $A=0$ and $C=I_2$ for the position, the extended $H_\infty$ filter \eqref{eq:hinf-mean}--\eqref{eq:hinf-cov} reduces to a linear filter for $\hat p$ with Riccati state $P$, given in \eqref{eq:hinf_est}--\eqref{eq:hinf_ric} in Appendix~\ref{app:quad}. The augmented state is $z = [\hat x,\hat y,\theta,v_x,v_y,\omega,P_{xx},P_{yy},P_{xy}]^\top \in \mathbb{R}^9$ with the filter dynamics and
\begin{equation}
\dot\theta = \omega,\quad
\dot v_x = -\tfrac{(u_1+u_2)\sin\theta}{m} + d_x,\quad
\dot v_y = \tfrac{(u_1+u_2)\cos\theta}{m}-g + d_y,\quad
\dot\omega = \tfrac{a(u_1-u_2)}{I}.
\label{eq:quad_true}
\end{equation}
The adversary is $[d_x,d_y,e_1,e_2]^\top$ with wind $|d_i|\le\bar d$ and $e\in\mathcal{E}(P)=\{e:\ |e_i|\le\kappa\sqrt{P_{ii}+r}\}$, and the controller picks $u\in[0,u^{\max}]^2$.

\textbf{Parameters:} We train the value function using DeepReach over $T=1$\,s a 6D value function without estimation (plain BRT) and two 9D estimation-aware ones ($\kappa=2$ and $\kappa=3$) for the failure set $\{\|p\|<0.5\}$; the filter parameters give the isotropic steady state $P_\infty\approx0.0231$. Parameters, domains and training settings are in Appendices~\ref{app:quad} and~\ref{app:training}.

\textbf{Rollout evaluation:} A goal-seeking PD controller acting on $\hat p$ drives the quadrotor past the obstacle, alone or with one of five filters: the plain BRT at the certainty-equivalent state, the plain BRT with a constant margin $\kappa\sqrt{P_\infty}$ (ss) or $\kappa\sqrt{0.075}$ (max, largest covariance of the domain), the estimation-aware BRT at $z$, and the belief CBF \cite{bcbf}.
%, whose risk-aware half-plane with risk level $\delta=1-\Phi(\kappa)$ is tightened by $\kappa\sqrt{\alpha^\top P\alpha}$, $\alpha=\hat p/\|\hat p\|$; it enforces a second-order exponential CBF (gains $(4,4)$) robust to the worst-case wind on the outer-loop acceleration command ($|a_i|\le4$\,m/s$^2$), which the PD controller's inner loop converts to thrusts.
We run each method with $\kappa=2$ and $\kappa=3$, so that all uncertainty-dependent methods use the same multiplier, for $2000$ rollouts each ($6$\,s, $\Delta t=0.01$\,s) toward the goal $(2,0)$ from $x_0\in[-2.2,-1.8]$, $y_0\in[-0.6,0.6]$, $\theta_0\in[-0.1,0.1]$, $v_0\in[-0.2,0.2]^2$, $\omega_0=0$, with $P_0=P_\infty I_2$, $\hat p_0\sim\mathcal{N}(p_0,P_0)$ and the worst-case wind $d=-\bar d\operatorname{sign}(\partial V_{\mathrm{plain}}/\partial v)$ at the true state; value-function filters engage when $V\le0.05$, and initial states and noise are shared by all methods.

\begin{figure}[!tb]
\captionsetup{font=footnotesize}
\centering
\includegraphics[width=0.85\linewidth]{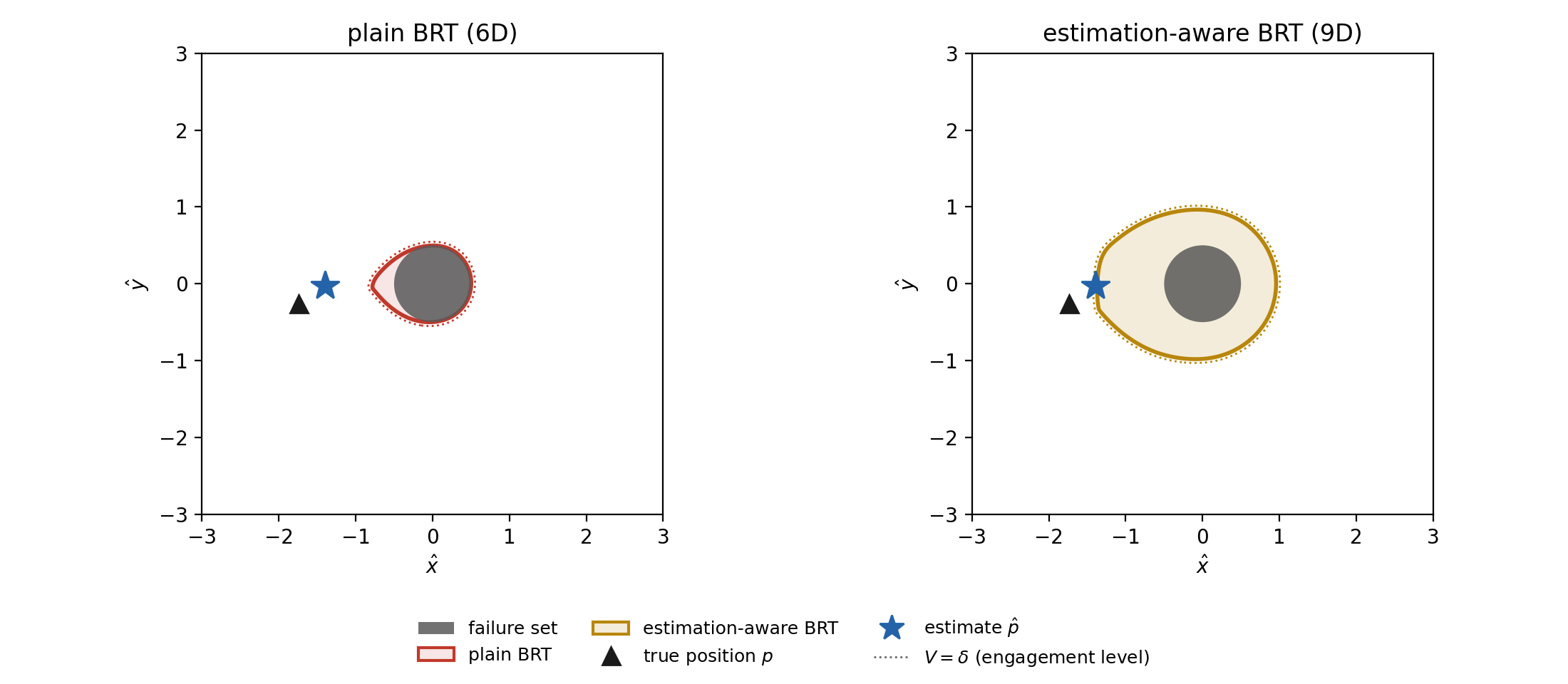}
\caption{A single sample rollout of the planar quadrotor with the plain and estimation-aware filters from the same initial state ($\kappa=3$). The BRT slices are taken at the time $t_e$ at which the estimation-aware filter first engages, at the same $(\theta,v_x,v_y,\omega)$. The estimation-aware unsafe set is larger than the plain one and already contains $\hat p$ at $t_e$, while the plain slice does not, so only the estimation-aware filter intervenes at this point.}
\label{fig:quad_brt_slice}
\end{figure}

\textbf{Results:} Fig.~\ref{fig:quad_brt_slice} illustrates the mechanism on one sample rollout: when the estimation-aware filter first engages, its unsafe set already contains $\hat p$ while the plain one does not; averaged over the $\kappa=3$ rollouts, it is $3.65\times$ larger in the $x$--$y$ plane at that state. In Table~\ref{tab:quad_kappa}, the plain filter collides in $70.8\%$ of rollouts because it engages late ($0.78$\,m from the obstacle center). The estimation-aware filter engages earlier ($1.22$--$1.41$\,m) and has the lowest intervention cost of all filters for both $\kappa$; it is collision-free for $\kappa=3$ and collides in $1.55\%$ of rollouts for $\kappa=2$, consistent with the bound of Theorem~\ref{thm:pointwise} dropping from $0.989$ to $0.865$. The maximum-margin baseline is also collision-free but more conservative, and the belief CBF collides in $2.5$--$16.7\%$ of rollouts at more than twice the cost.

\begin{table}[t]
\centering
\caption{Closed-loop results for the planar quadrotor ($n=2000$ rollouts per method and $\kappa$, same initial states, noise and wind). The $\kappa=2$ and $\kappa=3$ estimation-aware value functions are separately trained networks; the baselines use the same $\kappa$ in their margin or risk level. Clearance is $\min_t\|p\|-R_c$ and engagement distance is $\|p\|$ when the filter first acts, both with the true position. Margins: $\kappa\sqrt{P_\infty}$ (ss) and $\kappa\sqrt{0.075}$ (max). Collision rates with 95\% Wilson intervals.}
\label{tab:quad_kappa}
\resizebox{\linewidth}{!}{%
\begin{tabular}{llcccc}
\toprule
Method & $\kappa$ & \makecell{Collision rate [\%]\\(95\% CI)} & \makecell{Mean / worst\\clearance [m]} & \makecell{Engagement\\dist.\ [m]} & \makecell{Intervention\\$\int\|u-u_{\mathrm{nom}}\|dt$} \\
\midrule
Nominal only & --- & $100$ $[99.8,\,100]$ & $-0.441$ / $-0.500$ & --- & $0.00$ \\
Plain BRT & --- & $70.8$ $[68.7,\,72.7]$ & $-0.078$ / $-0.487$ & $0.78$ & $3.47$ \\
\midrule
\multirow{2}{*}{Plain + margin (ss)} & 2 & $4.20$ $[3.41,\,5.17]$ & $0.267$ / $-0.238$ & $1.13$ & $3.38$ \\
 & 3 & $0.20$ $[0.08,\,0.51]$ & $0.428$ / $-0.100$ & $1.29$ & $3.39$ \\
\midrule
\multirow{2}{*}{Plain + margin (max)} & 2 & $0$ $[0,\,0.19]$ & $0.519$ / $0.024$ & $1.38$ & $3.47$ \\
 & 3 & $0$ $[0,\,0.19]$ & $0.791$ / $0.340$ & $1.63$ & $3.61$ \\
\midrule
\multirow{2}{*}{Belief CBF} & 2 & $16.7$ $[15.1,\,18.4]$ & $0.130$ / $-0.341$ & $1.88$ & $7.77$ \\
 & 3 & $2.50$ $[1.90,\,3.28]$ & $0.279$ / $-0.230$ & $1.96$ & $7.98$ \\
 \midrule
 \multirow{2}{*}{Estimation-aware} & 2 & $1.55$ $[1.09,\,2.19]$ & $0.316$ / $-0.164$ & $1.22$ & $3.03$ \\
 & 3 & $0$ $[0,\,0.19]$ & $0.490$ / $0.032$ & $1.41$ & $2.93$ \\
\bottomrule
\end{tabular}%
}
\end{table}

\subsection{Dubins Car with Position-Dependent Measurement Noise}
\label{sec:dubins}

\noindent\textbf{Setup:} We now consider a case in which the full state is estimated and the information gained depends on the vehicle's position and heading. A Dubins car with state $x=[p_x,p_y,\theta]^\top$, constant speed $v$ and turn rate $|\omega|\le\omega_{\max}$ measures only its position, through a beacon at $b\in\mathbb{R}^2$ whose noise intensity $r(p)=r_0+r_1\|p-b\|^2$ grows with distance; the heading is never measured. The CT-EKF \eqref{eq:ctekf-mean}--\eqref{eq:ctekf-cov} then yields the 9D augmented state
\begin{equation}
z = [\hat x,\hat y,\hat\theta,\Sigma_{xx},\Sigma_{yy},\Sigma_{\theta\theta},\Sigma_{xy},\Sigma_{x\theta},\Sigma_{y\theta}]^\top \in \mathbb{R}^9 ,
\end{equation}
whose dynamics \eqref{eq:dubins_est}--\eqref{eq:dubins_cov} and innovation set \eqref{eq:dubins_innov} are given in Appendix~\ref{app:dubins}. The only adversary is the position innovation, which also corrupts $\hat\theta$ through the cross-covariances, and heading uncertainty inflates the position uncertainty along the direction of travel. Since every measurement term scales with $1/r(\hat p)$, information is gained quickly near the beacon and slowly far from it, so the trajectory shapes the covariance. The value function accounts for this effect, whereas a myopic filter acting on the current belief does not.

\noindent\textbf{Parameters:} We use $v=1$, $\omega_{\max}=1.2$, $R_c=0.5$, $\kappa=3$, a beacon at $b=(2,0)$ with $r_0=r_1=0.01$, and $q_x=q_y=0.005$, $q_\theta=0.01$. Over the domain, $r(\hat p)$ ranges from $0.01$ at the beacon to $0.35$ at the farthest corner, so the obstacle inflation $\kappa\sqrt{\lambda_{\max}(\Sigma_{pp})}$ varies from about $0.3$ near the beacon to about $1.2$ far from it; as in Air3D, Theorem~\ref{thm:pointwise} gives a bound of $0.989$. The domain and training settings are in Appendices~\ref{app:dubins} and~\ref{app:training}.

\begin{figure}[!tb]
\captionsetup{font=footnotesize}
\centering
\includegraphics[width=0.9\linewidth]{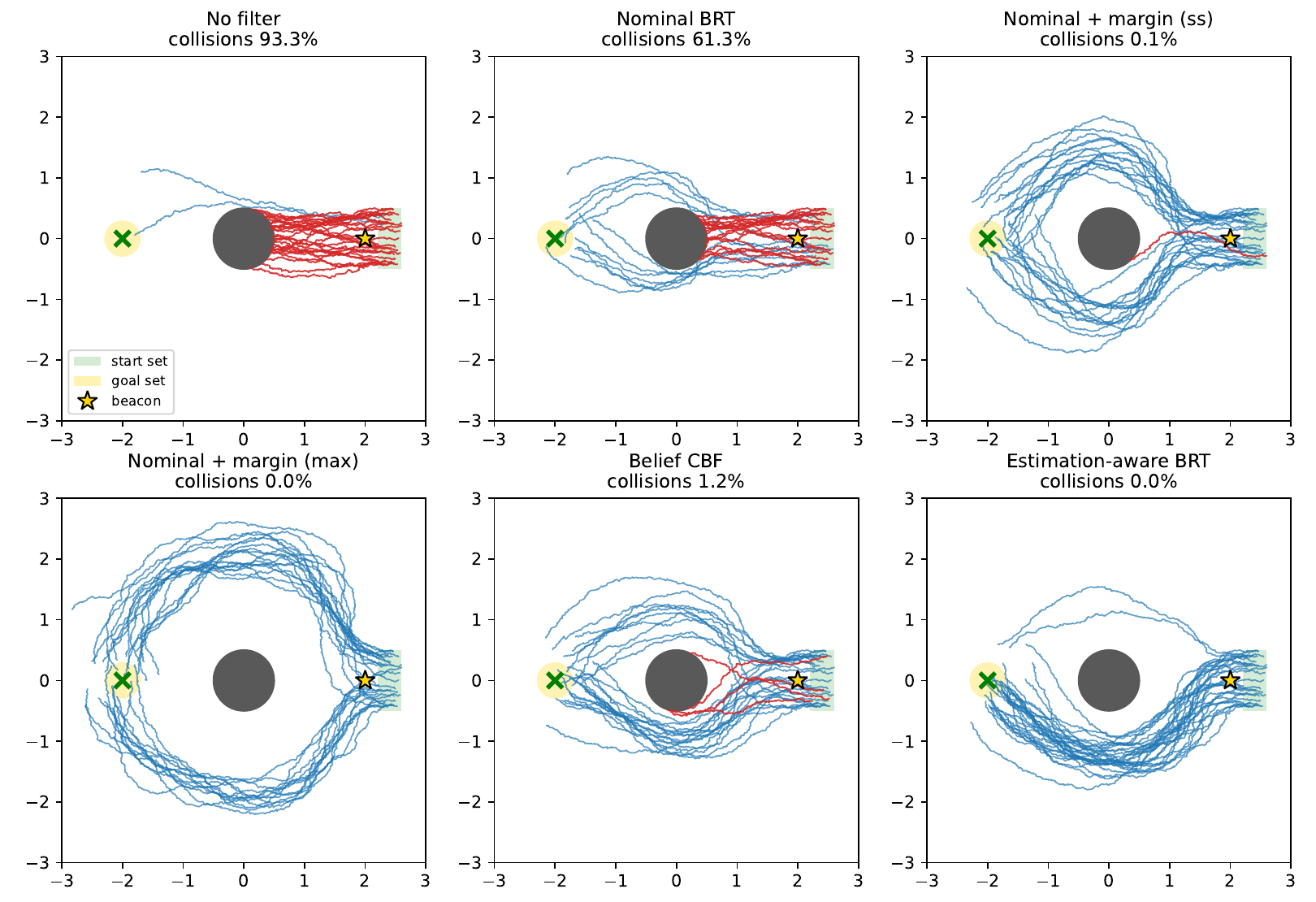}
\caption{Sample true trajectories of the Dubins car for each method (red: colliding rollouts), starting near the beacon (star) and driving to the goal (cross).}
\label{fig:dubins_traj}
\end{figure}

\noindent\textbf{Safety filter:} A heading controller acting on the estimate ($K_p = 2$) steers the car from near the beacon, $x_0\in[2.2,2.6]$, $y_0\in[-0.5,0.5]$, $\theta_0\in[\pi-0.1,\pi+0.1]$, through the obstacle toward the goal $(-2,0)$, away from the beacon. The CT-EKF starts at the steady-state covariance of the initial position (mean $\lambda_{\max}(\Sigma_{pp})=0.019$), so the uncertainty grows along the way. The car declares the goal reached when its estimate is within $0.3$\,m of it. We compare no filter; the nominal Dubins BRT (grid-based, \cite{hj_reach_ASL2023}) at the certainty-equivalent state; the nominal BRT with a constant margin $\kappa\sqrt{\bar\lambda}$, with $\bar\lambda$ the steady-state $\lambda_{\max}(\Sigma_{pp})$ at the obstacle ($0.67$) or the domain maximum ($1.47$); the estimation-aware BRT queried at $z$; and the belief CBF \cite{bcbf} of Section~\ref{sec:quad} acting on $\omega$. We simulate $2000$ rollouts per method ($8$\,s) with shared initial states and noise and no adversarial disturbance, so failures stem from the estimation error alone.

\begin{table}[t]
\centering
\caption{Dubins car safety filter ($n=2000$ rollouts per method, $\kappa=3$). Collision rates with 95\% Wilson intervals. Rollouts end at a collision, so the clearance $\min_t\|p\|-R_c$ of a colliding rollout is about zero. Engagement distance is the true $\|p\|$ when the filter first acts; active is the fraction of time steps in which the filter acts; goal is the fraction of rollouts whose estimate reaches the goal, and goal error the mean true distance to the goal at that moment.}
\label{tab:dubins}
\resizebox{\linewidth}{!}{%
\begin{tabular}{lccccccc}
\toprule
Method & \makecell{Collision rate [\%]\\(95\% CI)} & \makecell{Mean / worst\\clearance [m]} & \makecell{Engagement\\dist.\ [m]} & Active & \makecell{Intervention\\$\int|\omega-\omega_{\mathrm{nom}}|dt$} & Goal & \makecell{Goal\\error [m]} \\
\midrule
No filter & $93.4$ $[92.2,\,94.4]$ & $0.001$ / $-0.028$ & --- & --- & $0.00$ & $6.5\%$ & $0.49$ \\
Nominal BRT & $61.3$ $[59.1,\,63.4]$ & $0.043$ / $-0.029$ & $0.98$ & $0.19$ & $0.98$ & $38.7\%$ & $0.45$ \\
Nominal + margin (ss) & $0.15$ $[0.05,\,0.44]$ & $0.533$ / $-0.007$ & $1.79$ & $0.29$ & $2.71$ & $99.4\%$ & $0.41$ \\
Nominal + margin (max) & $\mathbf{0}$ $[0,\,0.19]$ & $1.136$ / $0.413$ & $2.40$ & $0.49$ & $5.25$ & $96.8\%$ & $0.42$ \\
Belief CBF & $1.20$ $[0.81,\,1.78]$ & $0.447$ / $-0.015$ & $1.76$ & $0.40$ & $2.19$ & $98.7\%$ & $0.41$ \\
Estimation-aware BRT & $\mathbf{0}$ $[0,\,0.19]$ & $0.695$ / $0.025$ & $2.11$ & $0.29$ & $3.34$ & $99.9\%$ & $0.41$ \\
\bottomrule
\end{tabular}%
}
\end{table}

\noindent\textbf{Results:} In Table~\ref{tab:dubins}, the nominal filter collides in $61.3\%$ of rollouts because it engages late, $1.0$\,m from the obstacle center. The steady-state margin still collides in $0.15\%$ of rollouts. The maximum margin and the estimation-aware filter are both collision-free, but the maximum margin engages immediately and acts in $49\%$ of time steps, whereas the estimation-aware filter engages only once the growing uncertainty requires it ($0.31$\,s), acts in $29\%$ of time steps, and cuts the intervention cost by $36\%$. The belief CBF collides in $1.2\%$ of rollouts, and its QP is infeasible in $13\%$ of time steps, when the car heads at the obstacle center and $\omega$ has no first-order effect on the constraint. Its constraint assumes zero innovation, so unmodeled measurement updates push the belief below its barrier in $78\%$ of rollouts even while the QP is feasible, and the constant-speed car collides when the true error points inward; the estimation-aware filter instead treats the innovation as a bounded adversary. All methods stop about $0.41$\,m from the true goal, consistent with the position uncertainty there (standard deviation up to $0.35$\,m).

\section{Conclusion}
\label{sec:conclusion}

This paper formulated a Hamilton--Jacobi reachability framework that is aware
of state estimation. Unlike standard HJ reachability, which assumes the state
is perfectly known, our formulation operates on a belief-augmented state
combining the perfectly known states, the estimator output, and the
covariance. The measurement innovation is confined to a
covariance-dependent set $\mathcal{E}(\Sigma)$ derived from the innovation
covariance, and the estimator's Riccati dynamics are integrated into the
Hamiltonian. Conservatism with
respect to estimation uncertainty is encoded in the target set, which is
inflated (or eroded) by an uncertainty margin scaled by the confidence
parameter $\kappa$, and every margin admits a pointwise probabilistic
interpretation: whenever the value function certifies safety, the true state
avoids the nominal unsafe region with probability at least $1 -
\delta_x(\kappa)$, provided the estimator is consistent.

The framework is computationally practical: beyond the belief-augmented state, it modifies only the Hamiltonian and the target set, allowing existing HJ reachability computation pipelines to
be adapted with minimal changes, and for
the value and gradients to be used as part of a safety filter whose
behavior automatically tightens or relaxes with the estimator's current
uncertainty. The framework provides a pointwise probabilistic guarantee on the true state,
conditional on a consistent Gaussian estimator; extending it to a
trajectory-level certificate, which would turn the empirical closed-loop safety
statistics reported here into certified ones, and reducing the conservatism of
worst-case innovation modeling are natural next steps. The effectiveness of the framework was demonstrated through Air3D pursuit-evasion and fully estimated Dubins car and planar-quadrotor case studies, where the estimation-aware filters substantially reduced collision rates compared with certainty-equivalent and other baseline approaches, while adapting their conservatism to the estimator's evolving uncertainty.

\newpage
\bibliographystyle{unsrt}
\bibliography{references}

\appendix
\section{DeepReach training settings}
\label{app:training}

All value functions are learned by defining the augmented systems using the DeepReach implementation \cite{deepreach2024} over the  horizon $T=1$\,s: after pretraining on the terminal value, the HJI loss is supplemented with MPC supervision \cite{MPCAD}. Table~\ref{tab:hyper} lists the settings; system-specific domains are given in the following sections.

\begin{table}[H]
\centering
\caption{DeepReach training settings. All networks use a sine-activated MLP \cite{siren} with the exact DeepReach formulation \cite{exact}, Adam, and horizon $T=1$\,s. Iterations include the $10{,}000$ pretraining iterations.}
\label{tab:hyper}
\resizebox{\linewidth}{!}{%
\begin{tabular}{lcccc}
\toprule
 & \makecell{Air3D\\(6D)} & \makecell{Quadrotor\\plain (6D)} & \makecell{Quadrotor\\est.-aware (9D)} & \makecell{Dubins\\est.-aware (9D)} \\
\midrule
Hidden layers $\times$ units & $3\times512$ & $3\times512$ & $3\times512$ & $3\times512$ \\
Learning rate & $2\times10^{-5}$ & $2\times10^{-5}$ & $2\times10^{-5}$ & $2\times10^{-5}$ \\
Pretraining iterations & $10{,}000$ & $10{,}000$ & $10{,}000$ & $10{,}000$ \\
Training iterations & $120{,}000$ & $100{,}000$ & $150{,}000$ & $150{,}000$ \\
Curriculum ends at iteration & $100{,}000$ & $90{,}000$ & $135{,}000$ & $135{,}000$ \\
Collocation points / iteration & $65{,}000$ & $65{,}000$ & $65{,}000$ & $65{,}000$ \\
MPC dataset size & $10{,}000$ & $20{,}000$ & $30{,}000$ & $30{,}000$ \\
MPC batches / step ($\times1024$) & $20$ & $20$ & $30$ & $30$ \\
Perturbations per MPC sample & $100$ & $100$ & $100$ & $100$ \\
MPC refinement rounds & $10$ & $10$ & $10$ & $20$ \\
\bottomrule
\end{tabular}%
}
\end{table}

\section{Air3D: additional details}
\label{app:air3d}

\noindent\textbf{Covariance dynamics:} With $A = \begin{bmatrix} 0 & u_E \\ -u_E & 0 \end{bmatrix}$, $C = I_2$, $R = rI_2$ and $Q = qI_2$, the CT-EKF Riccati equation \eqref{eq:ctekf-cov} reads
\begin{equation}
\begin{aligned}
\label{eq:covariance}
\dot{\Sigma}_{xx} &= 2u_E\Sigma_{xy} + q - \tfrac{1}{r}\left(\Sigma_{xx}^2 + \Sigma_{xy}^2\right), \\
\dot{\Sigma}_{yy} &= -2u_E\Sigma_{xy} + q - \tfrac{1}{r}\left(\Sigma_{xy}^2 + \Sigma_{yy}^2\right), \\
\dot{\Sigma}_{xy} &= u_E(\Sigma_{yy} - \Sigma_{xx}) - \tfrac{1}{r}\Sigma_{xy}(\Sigma_{xx} + \Sigma_{yy}).
\end{aligned}
\end{equation}

\noindent\textbf{Noise levels:} On the isotropic slice $\Sigma$ converges to $\sqrt{qr}=0.05$, where $K\approx I_2$ and $|e_i|\le3\sqrt{0.1}\approx0.95$, so the adversarial term $Ke$ is comparable to the vehicle speeds; at $\sigma=0.001$, $K\approx0.02\,I_2$ and the innovation has almost no effect. Smaller noise would yield a BRT closer to the nominal one.

\noindent\textbf{Training:} The value function is learned on the domain given in Section~\ref{sec:air3d} with MPC-guided adversarial refinement \cite{MPCAD} (settings in Appendix~\ref{app:training}; about 1.8 hours on a single GPU).

\noindent\textbf{Rollouts:} The evader's nominal command is a saturated proportional heading law (gain $3$). The pursuer plays $u_P=-u_P^{\max}\operatorname{sign}(\partial V_{\mathrm{std}}/\partial\psi)$ of the nominal Air3D value function at the true relative state; the true system has process noise $q$, the measurements noise $r$, and the estimate is propagated by the CT-EKF. Initial headings are $\psi_0=\pi/2$ and the horizon is $5$\,s. For the avoid problem the constant margin is exactly the threshold shift $V-\kappa\sqrt{\bar\sigma}\le0.05$ (margins $0.67$ and $0.82$). The belief CBF uses the barrier $\|\hat p\|-R_c-\kappa\sqrt{\alpha^\top\Sigma\alpha}$, $\alpha=\hat p/\|\hat p\|$, with a second-order exponential CBF constraint (gains $(4,4)$) that holds for every $|u_P|\le3$, solved as a QP at every step.

\subsection{Comparison with grid-based solutions}
\label{app:air3d_grid}

\begin{figure}[H]
  \centering
  \includegraphics[width=\linewidth]{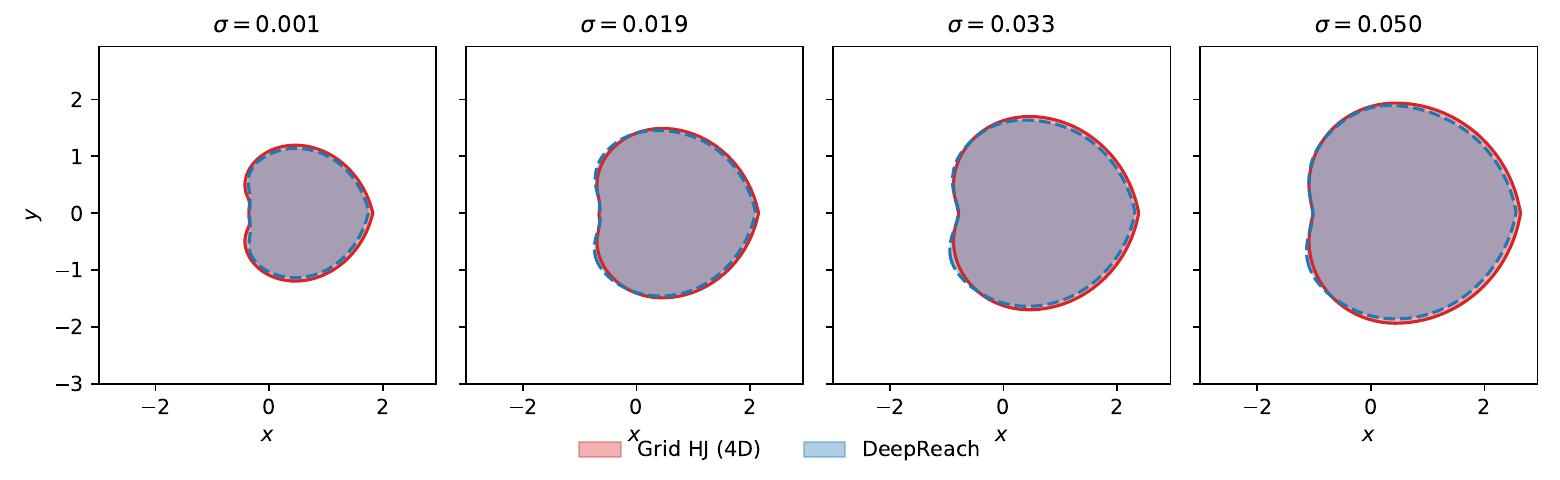}
  \caption{Learned (DeepReach) and \texttt{hj\_reachability} grid estimation-aware BRTs at $\psi=\pi$.}
  \label{fig:grid_vs_dr}
\end{figure}

\noindent\textbf{Learned vs.\ grid BRT:} The isotropic slice is invariant under \eqref{eq:covariance} ($\dot\Sigma_{xy}=0$, $\dot s = q - s^2/r$ for any $u_E$), so on it the 6D problem is exactly a 4D problem in $(\hat x,\hat y,\psi,s)$. We solve it with \texttt{hj\_reachability} \cite{hj_reach_ASL2023} (\texttt{very\_high} accuracy) on a $121\times121\times48\times15$ grid over $\hat x,\hat y\in[-4,4]$, cropped to the learned domain for the comparison. The learned and grid BRTs agree with IoU $0.95$ for $\sigma\ge0.019$ and $0.92$ at $\sigma=0.001$ (Table~\ref{tab:grid_vs_dr}, Fig.~\ref{fig:grid_vs_dr}). The learned BRT slightly under-approximates the grid BRT ($4$--$5\%$ of it is marked safe for $\sigma\ge0.019$, $8.5\%$ at $\sigma=0.001$). This validates the learned value function on the isotropic slice only, not on the anisotropic part of the domain.

\begin{table}[t]
  \centering
  \caption{Learned vs.\ grid-based BRT at $t=T$ on the slice $\Sigma_{xx}=\Sigma_{yy}=\sigma$, $\Sigma_{xy}=0$.
  IoU, miss rate and Hausdorff distance (HD) are averaged over
  $\psi\in\{-\pi,-\pi/2,0,\pi/2,\pi\}$. Volume error (\% of the $(\hat x,\hat y,\psi)$ domain) and value MSE
  are computed over the full grid at each $\sigma$.}
  \label{tab:grid_vs_dr}
  \begin{tabular}{lccccc}
    \toprule
    $\sigma$ & IoU & Miss rate (\%) & HD (cells) & Vol.\ error (\%) & MSE \\
    \midrule
    0.001 & 0.915 & 8.5 & 1.4 & 1.05 & $8.3\times10^{-3}$ \\
    0.019 & 0.952 & 3.9 & 1.3 & 0.83 & $2.4\times10^{-3}$ \\
    0.033 & 0.948 & 4.4 & 1.6 & 1.22 & $3.1\times10^{-3}$ \\
    0.050 & 0.952 & 4.5 & 1.6 & 1.47 & $3.6\times10^{-3}$ \\
    \bottomrule
  \end{tabular}
\end{table}

\begin{figure}[H]
    \centering
    \includegraphics[width=\linewidth]{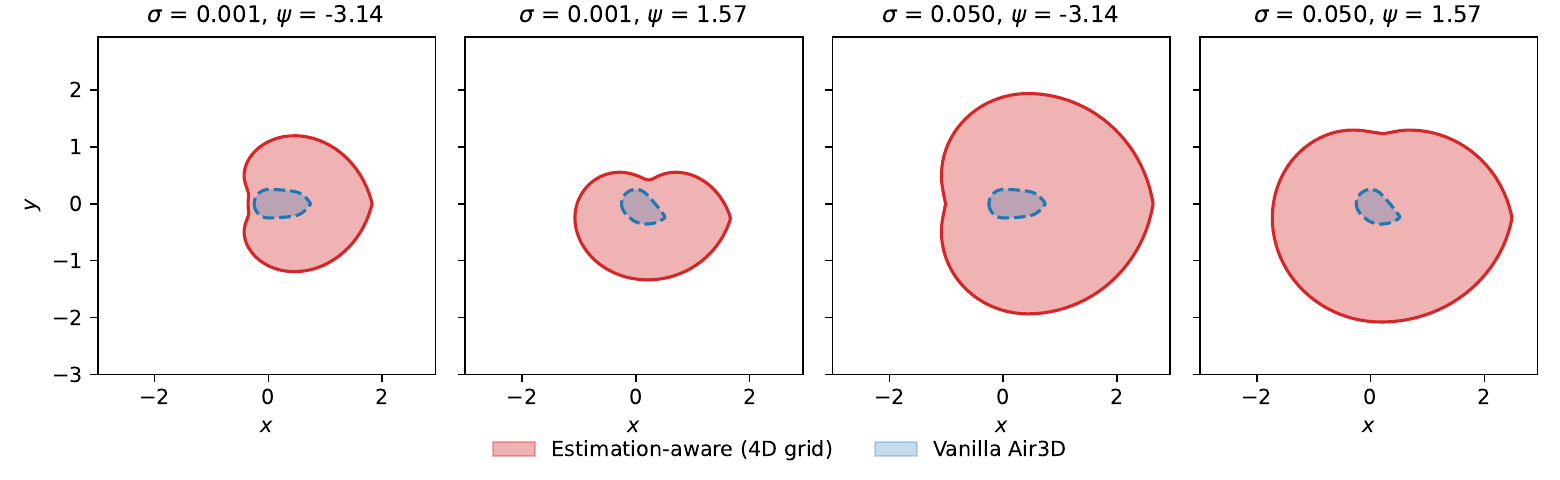}
    \caption{Nominal Air3D BRT (perfect state knowledge) and grid-based estimation-aware BRT for initial covariance levels $0.001$ and $0.05$.}
    \label{fig:nominal_vs_estaware_0p001}
\end{figure}

\noindent\textbf{Nominal vs.\ estimation-aware BRT:} Fig.~\ref{fig:nominal_vs_estaware_0p001} compares the grid estimation-aware BRT with the nominal 3D Air3D BRT computed with the same toolbox and settings. Even for $\Sigma_0=0.001$ the two differ, because with $q\neq0$ the covariance grows quickly ; the nominal BRT is recovered only when $Q=0$ as well (Remark in Section~\ref{sec:method}). At $t=T$ the estimation-aware BRT contains the nominal one on every slice we evaluated, and its $(\hat x,\hat y)$ area is on average $13.6\times$ larger at $\sigma=0.001$ and $37\times$ larger at $\sigma=0.05$.

\section{Planar quadrotor: additional details}
\label{app:quad}

\noindent\textbf{$H_\infty$ filter:} The filter models the position as $\dot p = v + Bw$, $y_m = p + Dn$, with $BB^\top = qI_2$, $DD^\top = rI_2$, so $A=0$ and $C=L_m=I_2$, and the extended $H_\infty$ filter \eqref{eq:hinf-mean}--\eqref{eq:hinf-cov} reduces to the linear filter
\begin{align}
\dot{\hat p} &= v + \tfrac{1}{r}P\,e, \qquad e = y_m-\hat p, \label{eq:hinf_est}\\
\dot P &= qI_2 - \Big(\tfrac{1}{r}-\tfrac{1}{\gamma^2}\Big)P^2, \label{eq:hinf_ric}
\end{align}
with gain $K=PC^\top(DD^\top)^{-1}=P/r$. Writing $c=1/r-1/\gamma^2$, the three Riccati states of the augmented state $z$ evolve as
\begin{equation}
\dot P_{xx} = q - c\,(P_{xx}^2+P_{xy}^2),\quad
\dot P_{yy} = q - c\,(P_{yy}^2+P_{xy}^2),\quad
\dot P_{xy} = -c\,P_{xy}(P_{xx}+P_{yy}).
\label{eq:hinf_ric_comp}
\end{equation}
In the Hamiltonian, the innovation $e$ in \eqref{eq:hinf_est} is the adversarial input confined to $\mathcal{E}(P)$, while \eqref{eq:hinf_ric_comp} does not depend on $e$ or on the vehicle state.

\noindent\textbf{Parameters and training:} We use $u^{\max}=8$, $\bar d=0.5$, $q=0.01$, $r=0.04$, $\gamma=0.4$ and $\kappa\in\{2,3\}$. These satisfy $1/r>1/\gamma^2$, and the isotropic steady state is $P_\infty=\sqrt{q/(1/r-1/\gamma^2)}\approx0.0231$. As in Air3D, the noise is intentionally large relative to the obstacle so that the effect of the estimator is visible. All three value functions are learned on $x,y\in[-3,3]$\,m, $|\theta|\le\pi/3$, $|v_x|,|v_y|\le2$\,m/s, $|\omega|\le4$\,rad/s, and the 9D ones also on $P_{xx},P_{yy}\in[0.001,0.05]$, $|P_{xy}|\le0.025$ (for $\kappa=2$, $x,y\in[-2.5,2.5]$\,m). After pretraining on the terminal value, the HJI loss is supplemented with MPC supervision \cite{MPCAD} (step $0.02$\,s, $10$ refinement rounds every $0.2$\,s). Training settings are in Appendix~\ref{app:training}.

\section{Dubins car: additional details}
\label{app:dubins}

\noindent\textbf{System and measurement model:} The Dubins car with state $x=[p_x,p_y,\theta]^\top$, constant speed $v$ and turn-rate input $\omega$ evolves as
\begin{equation}
\dot x = g(x,\omega) + w = \begin{bmatrix} v\cos\theta \\ v\sin\theta \\ \omega \end{bmatrix} + w,
\qquad |\omega|\le\omega_{\max},
\label{eq:dubins_true}
\end{equation}
with process noise of intensity $Q=\mathrm{diag}(q_x,q_y,q_\theta)$. Only the position is measured, by a beacon at $b\in\mathbb{R}^2$ whose noise grows with distance:
\begin{equation}
y_m = Cx + n, \qquad C = \begin{bmatrix} I_2 & 0 \end{bmatrix}, \qquad
R(p) = r(p)\,I_2, \qquad r(p) = r_0 + r_1\|p-b\|^2 .
\label{eq:dubins_meas}
\end{equation}
The heading is never measured directly. The CT-EKF \eqref{eq:ctekf-mean}--\eqref{eq:ctekf-cov} uses
\begin{equation}
A(\hat\theta) = \left.\frac{\partial g}{\partial x}\right|_{\hat x} =
\begin{bmatrix} 0 & 0 & -v\sin\hat\theta \\ 0 & 0 & v\cos\hat\theta \\ 0 & 0 & 0 \end{bmatrix},
\qquad
K = \frac{1}{r(\hat p)}\,\Sigma C^\top \in \mathbb{R}^{3\times2},
\label{eq:dubins_jac}
\end{equation}
where the filter evaluates the noise level at the estimated position $\hat p$, since the true position is unknown. With the augmented state $z$ of Section~\ref{sec:dubins}, innovation $e=[e_1,e_2]^\top$ and $r \coloneqq r(\hat p)$, the estimate evolves as
\begin{equation}
\begin{aligned}
\dot{\hat x} &= v\cos\hat\theta + \tfrac{1}{r}\left(\Sigma_{xx}e_1 + \Sigma_{xy}e_2\right), \\
\dot{\hat y} &= v\sin\hat\theta + \tfrac{1}{r}\left(\Sigma_{xy}e_1 + \Sigma_{yy}e_2\right), \\
\dot{\hat\theta} &= \omega + \tfrac{1}{r}\left(\Sigma_{x\theta}e_1 + \Sigma_{y\theta}e_2\right),
\end{aligned}
\label{eq:dubins_est}
\end{equation}
and, writing $s=\sin\hat\theta$ and $c=\cos\hat\theta$, the covariance \eqref{eq:ctekf-cov} becomes
\begin{equation}
\begin{aligned}
\dot\Sigma_{xx} &= -2vs\,\Sigma_{x\theta} + q_x - \tfrac{1}{r}\left(\Sigma_{xx}^2+\Sigma_{xy}^2\right), \\
\dot\Sigma_{yy} &= 2vc\,\Sigma_{y\theta} + q_y - \tfrac{1}{r}\left(\Sigma_{xy}^2+\Sigma_{yy}^2\right), \\
\dot\Sigma_{\theta\theta} &= q_\theta - \tfrac{1}{r}\left(\Sigma_{x\theta}^2+\Sigma_{y\theta}^2\right), \\
\dot\Sigma_{xy} &= v\left(c\,\Sigma_{x\theta} - s\,\Sigma_{y\theta}\right) - \tfrac{1}{r}\Sigma_{xy}\left(\Sigma_{xx}+\Sigma_{yy}\right), \\
\dot\Sigma_{x\theta} &= -vs\,\Sigma_{\theta\theta} - \tfrac{1}{r}\left(\Sigma_{xx}\Sigma_{x\theta}+\Sigma_{xy}\Sigma_{y\theta}\right), \\
\dot\Sigma_{y\theta} &= vc\,\Sigma_{\theta\theta} - \tfrac{1}{r}\left(\Sigma_{xy}\Sigma_{x\theta}+\Sigma_{yy}\Sigma_{y\theta}\right).
\end{aligned}
\label{eq:dubins_cov}
\end{equation}
The innovation is confined to $e\in\mathcal{E}(\Sigma,\hat p)$. With $S = C\Sigma C^\top + r(\hat p) I_2$, the set $\mathcal{E}$ defined in Section~\ref{sec:method} becomes
\begin{equation}
\mathcal{E}(\Sigma,\hat p) = \left\{ e\in\mathbb{R}^2 \;\middle|\; |e_1| \le \kappa\sqrt{\Sigma_{xx} + r(\hat p)},\ |e_2| \le \kappa\sqrt{\Sigma_{yy} + r(\hat p)} \right\},
\label{eq:dubins_innov}
\end{equation}
% which now depends on the estimated position as well as on $\Sigma$. The innovation $e=y_m-C\hat x$ lives in the measurement space, so it has only two components because only position is measured. The heading estimate is still perturbed by it, through the third row of $K$ in \eqref{eq:dubins_est}: an adversarial position innovation also corrupts $\hat\theta$ in proportion to $\Sigma_{x\theta}$ and $\Sigma_{y\theta}$.

\noindent\textbf{Training domain:} The value function is learned over $t\in[0,1]$ on $\hat x,\hat y\in[-3,3]$, $\hat\theta\in[-\pi,\pi]$, $\Sigma_{xx},\Sigma_{yy}\in[0.001,0.18]$, $\Sigma_{\theta\theta}\in[0.001,0.05]$, $|\Sigma_{xy}|,|\Sigma_{x\theta}|,|\Sigma_{y\theta}|\le0.06$. These ranges cover the 1st to 99th percentiles of the covariance along closed-loop simulations of \eqref{eq:dubins_true}--\eqref{eq:dubins_meas} with the CT-EKF inside the domain.

\noindent\textbf{Rollouts:} The nominal BRT is computed with \texttt{hj\_reachability} \cite{hj_reach_ASL2023} on a $161\times161\times120$ grid. Value-function filters engage when $V\le0.05$; the belief CBF uses gains $(4,4)$ and rollouts use $\Delta t=0.01$\,s.
\end{document}